**Type of Presentation**:  **Topic:** Deep Learning
Oral: ☒   In-person: ☒
Poster: ☐   Virtual in Zoom: ☐
The same: ☐   Pre-recorded video: ☐

# Porting Large Language Models to Mobile Devices for Question Answering

**Hannes Fassold** [1]
[1] JOANNEUM RESEARCH – DIGITAL, Steyrergasse 17, 8010 Graz
Tel.: +43 316 876-1126, e-mail: hannes.fassold@joanneum.at

**Summary:** Deploying Large Language Models (LLMs) on mobile devices makes all the capabilities of natural language processing available on the device. An important use case of LLMs is question answering, which can provide accurate and contextually relevant answers to a wide array of user queries. We describe how we managed to port state of the art LLMs to mobile devices, enabling them to operate natively on the device. We employ the llama.cpp framework, a flexible and self-contained C++ framework for LLM inference. We selected a 6-bit quantized version of the Orca-Mini-3B model with 3 billion parameters and present the correct prompt format for this model. Experimental results show that LLM inference runs in interactive speed on a Galaxy S21 smartphone and that the model delivers high-quality answers to user queries related to questions from different subjects like politics, geography or history.

**Keywords:** deep learning, large language models, question answering, mobile devices, termux

## 1. Introduction

Large Language Models (LLMs) [1] on mobile devices enhance natural language processing and enable more intuitive interactions. These models empower applications such as advanced virtual assistants, language translation, text summarization or the extraction of key terms in the text (named entity extraction).

An important use case of LLMs is also question answering, which can provide accurate and contextually relevant answers to a wide array of user queries. For example, it can be used for fake news detection on a smartphone by querying the LLM about the validity of dubious claims made in a news article.

Because of the limited processing power of a typical smartphone, usually the queries for an LLM on a mobile device are processed in the cloud and the LLM output is sent back to the device. This is the standard workflow for the ChatGPT app and most other LLM-powered chat apps. But often this is not possible or desired, for example for journalists operating in areas with limited connectivity or under strict monitoring and surveillance of internet traffic (e.g. in authoritarian regimes). In this case, the processing has to be done on the device.

In this work, we therefore demonstrate how to port LLMs efficiently to mobile devices so that they run natively and in interactive speed on a mobile device. In the following section, we will describe the software framework we employ for running LLMs natively on a mobile device (like a smartphone or tablet). Section 3 describes the specific model we have chosen and the proper prompt format for it. In section 4 we provide information about qualitative experiments with the LLM and section 5 concludes the paper.

## 2. LLM framework for on-device inference

Initially, we tried to do LLM inference natively on a mobile device via the *Tensorflow Lite* (TFLite) framework, as it is the most popular framework for on-device inference. But for LLMs, practically all finetuned models available on the *Hugging Face model hub* [1] provide only PyTorch weights, so a conversion to TFLite has to be done. Unfortunately, during our experiments we noticed that the the conversion pipeline is quite complex (PyTorch => ONNX => Tensorflow => TFLite) and not future-proof, as legacy Tensorflow 1.X versions have to be used in combination with code from several public github repos which are not maintained anymore.

We therefore opted for the *llama.cpp* framework [2], a very flexible and self-contained C++ framework for LLM inference on a variety of devices. It can run state of the art models like *Llama* / *Llama2* [2], *Vicuna* [3] or *Mistral* [4] either on CPU or GPU/CUDA and provides a lot of options (e.g. for setting temperature, context size or sampling method). It supports a variety of sub-8bit quantisation methods (from 2 bits to 6 bits per parameter), which is crucial for running models with billions of parameters on a smartphone with limited memory.

In order to build the C++ libraries and executables of the llama.cpp framework, a standard Linux build toolchain is needed consisting of a terminal (shell), command-line tools, CMake/Make, C/C++ compiler and linker and more. For *Android*, fortunately such a build toolchain is available via the *Termux* app [3]. It can be installed via the Android open-source software package manager F-Droid [4] and does not need root access to the device. It has been used already for deep learning tasks, for example for on-device training of neural networks as described in [5].

---

[1] https://huggingface.co/docs/hub/models-the-hub

[2] https://github.com/ggerganov/llama.cpp

After installation of Termux, we open a console there and install the necessary tools (like wget, git, cmake, clang compiler) via *pkg*, the Termux package manager. We install also the Android screen mirror software scrcpy [5] on the PC so that we can control the device directly on the PC and mirror its screen there.

For building the llama.cpp binaries, we now clone its latest sources from the respective github repo. We invoke *CMake* to generate the Makefile and build all binaries via the *make* command. We compile the binaries with model inference done on the CPU, as GPU inference relies on CUDA which is not available on Android devices. After compiling, several binaries are available on the device. The most important ones are an executable for direct interactive chatting with the LLM and a server application with an REST-API which is similar to the OpenAI API for ChatGPT. The server application allows for further integration, for example into a on-device GUI app for question answering.

## 3. Model selection and prompt format

On the Hugging Face model hub there are many pretrained large language models available, which differ in their network architecture, model size, training / finetuning procedure and dataset and their task (base model for text completion versus instruct model for instruction following and chat). After some experiments, we selected the *Orca-Mini-3B* model [6] with 3 billion parameters. It runs in interactive speed on a recent smartphone and provides decent responses to a user query due to finetuning via imitation learning with the Orca method [6]. We employ a quantized model with approximately 5.6 bit per parameter, which takes roughly 2.2 GB of CPU RAM on the device.

For an instruct model, it is important for a good performance of the model to use the same prompt format (system prompt, user prompt etc.) as was used for finetuning the model. For the Orca-Mini-3B model, this means that the prompt format has to be as shown in the following example:

*"### System:\n You are an AI assistant that follows instruction extremely well. Help as much as you can.\n\n### User:\n What is the smallest state in India ?\n\n### Response:\n"*

## 4. Experiments and evaluation

We did a subjective evaluation of the selected model by testing its responses for user queries related to questions from different subjects like politics, geography, history and more. From the tests we can infer that the model provides accurate and faithful answers for most of the user queries. Of course, like every LLM it can hallucinate (provide false information) from time to time.

An example output of the LLM application for direct chat can be seen in **Fig. 1**. The LLM provides correct answers for questions from different domains. The output of the model is generated fast enough for an interactive chat on a Samsung Galaxy S21 smartphone.

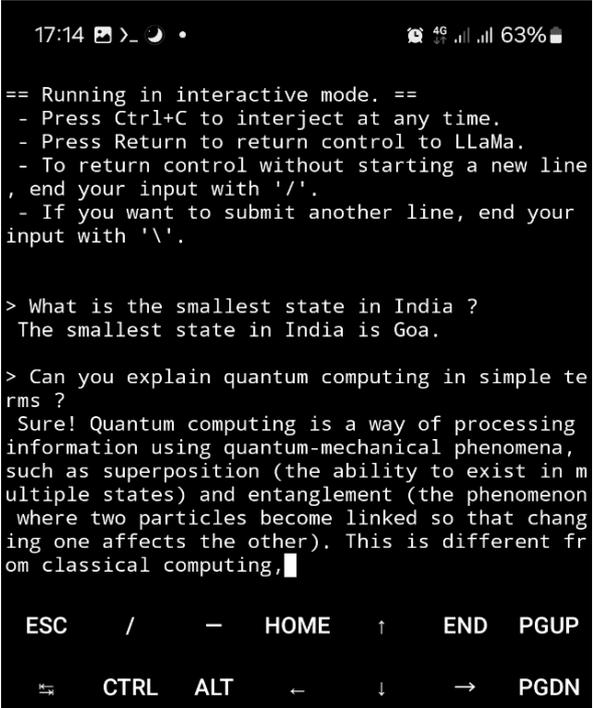

**Fig. 1.** Interactive chat application on a smartphone.

## 5. Conclusion

We demonstrated how to port large language models (LLMs) efficiently to mobile devices, so that they run natively on the device. We provided information on the LLM framework we employ as well as the model and proper prompt format for question answering. Experiments show that the model runs in interactive speed on a Galaxy S21 smartphone and provides high-quality answers for the user queries. In the future, we will explore recently introduced LLMs like *phi-2* [7] and GPU acceleration of the model via OpenCL or Vulkan on the device.

---

[3] https://termux.dev/en/
[4] https://f-droid.org/
[5] https://github.com/Genymobile/scrcpy
[6] https://huggingface.co/pankajmathur/orca_mini_3b

**Acknowledgements**

The research leading to these results has received funding from the European Union's Horizon 2020 research and innovation programme under grant agreement No. 951911 - AI4Media.

**References**

[1]. S. Minaee, T. Mikolov et al. Large Language Models: A Survey, arXiv:2402.06196, 2024.
[2]. H. Touvron, L. Martin. Llama 2: Open Foundation and Fine-Tuned Chat Models, arXiv:2307.09288, 2023.
[3]. L. Zheng, W. Chiang et al. Judging LLM-as-a-Judge with MT-Bench and Chatbot Arena, NeurIPS, 2023.
[4]. A. Jiang, A. Sablayrolles et al. Mistral 7B. arXiv:2310.06825, 2023.
[5]. S. Singapuram, F. Lai et al. Swan: A Neural Engine for Efficient DNN Training on Smartphone SoCs, IEEE Working Conference on Mining Software Repositories, 2022.
[6]. S. Mukherjee, A. Mitra et al. Orca: Progressive Learning from Complex Explanation Traces of GPT-4. arXiv:2306.02707, 2023.
[7]. M. Javaheripi, S. Bubeck et al. Phi-2: The surprising power of small language models, URL: https://www.microsoft.com/en-us/research/blog/phi-2-the-surprising-power-of-small-language-models/, 2023.